\documentclass[letterpaper,11pt]{article}

\usepackage{amssymb}
\usepackage{amsmath}
\usepackage{amsthm}
\usepackage{graphicx}
\usepackage{authblk}
\usepackage{float}
\usepackage{booktabs,siunitx}
\usepackage{multirow}
\usepackage{fullpage}
\usepackage{bm}
\usepackage[utf8x]{inputenc}
\usepackage{color}
\usepackage{subcaption}
\usepackage{subcaption}
\usepackage{algpseudocode, algorithm, algorithmicx}

\usepackage{setspace}

\title{Nonlinear Bayesian Update via Ensemble Kernel Regression with Clustering and Subsampling}
\author{Yoonsang Lee}
\date{}

\begin{document}
\maketitle
\begin{abstract}
Nonlinear Bayesian update for a prior ensemble is proposed to extend traditional ensemble Kalman filtering to settings characterized by non-Gaussian priors and nonlinear measurement operators. In this framework, the observed component is first denoised via a standard Kalman update, while the unobserved component is estimated using a nonlinear regression approach based on kernel density estimation. The method incorporates a subsampling strategy to ensure stability and, when necessary, employs unsupervised clustering to refine the conditional estimate. Numerical experiments on Lorenz systems and a PDE-constrained inverse problem illustrate that the proposed nonlinear update can reduce estimation errors compared to standard linear updates, especially in highly nonlinear scenarios.
\end{abstract}

%
%
%
%
%
%
%
%
%
\section{Introduction}
Ensemble-based methods play a crucial role in Bayesian inference by representing the probability distribution with a collection of samples that capture the inherent uncertainty of complex, nonlinear models. These methods enable efficient uncertainty propagation and sequential updating as new measurements become available, thereby facilitating robust Bayesian updates even in high-dimensional settings. This approach has proven valuable in numerous applications, including numerical weather prediction \cite{Evensen2009,ReichCotter2015}, robotics \cite{fox2001particle}, and neural network training \cite{wang1992extended, kovachki2019ensemble, fan2018gaussian}.

Traditional ensemble Kalman filters assume Gaussian statistics for both the state and its errors. While this assumption simplifies the update, it can be overly restrictive for systems exhibiting strong nonlinearity or non-Gaussian behavior \cite{Snyder2008}. Although particle filters \cite{PF} can alleviate some of these limitations, they often suffer from particle degeneracy and require significantly larger ensembles. Consequently, ensemble-based Kalman filters remain popular in high-dimensional settings, particularly when enhanced by techniques such as covariance inflation and localization \cite{loc2, Anderson2007, Hoteit1}.

In this paper, we revisit the interpretation of the Bayesian update as a regression problem linking the observed and unobserved components of the state. This perspective emphasizes two key challenges that can affect filter performance: i) accurately capturing nonlinear relationships between observations and state estimates, and ii) addressing the inherent nonuniqueness in this mapping. Our approach decomposes the update into two steps. First, the observed components are denoised using standard ensemble-based filtering techniques. Next, the unobserved components are estimated via regression—potentially employing nonlinear, nonparametric models—by leveraging their relationship with the observed variables.

Our method employs kernel density estimation (KDE) to perform nonlinear regression on the unobserved state given the observed state. KDE’s flexibility in capturing non-Gaussian features makes it a natural choice that can be further refined or extended with alternative nonparametric approaches in future work. In contrast, Gaussian mixture model–based methods \cite{govaers2015gaussian, reich2012gaussian} approximate the posterior as a weighted sum of Gaussian components, which enables them to capture multimodality but requires careful selection of the number of mixtures and tuning of their parameters. The proposed nonlinear Bayesian update directly estimates the conditional mean without explicitly resolving mixture weights or covariances, thereby avoiding these complexities and reducing computational overhead. Furthermore, by incorporating local subsampling and clustering techniques, our approach demonstrates enhanced robustness in scenarios with limited ensemble sizes or strong nonlinear correlations, offering a practical alternative to Gaussian mixture–based inference schemes.

The remainder of the paper is organized as follows. Section \ref{sec:review} outlines the state and measurement models and reviews the standard Kalman update, which is interpreted as a linear regression between observed and unobserved components. Building on this framework, Section \ref{sec:NlBU} introduces the nonlinear Bayesian update that employs nonlinear regression, and discusses strategies such as subsampling and clustering to enhance robustness. Finally, Section \ref{sec:experiment} presents numerical experiments involving Lorenz models and a PDE-constrained inverse problem to demonstrate the effectiveness of the proposed method.

%
%
%
%
%
%
%
%
%
\section{Standard Bayesian Update}\label{sec:review}
In this section, we formulate the state variable and measurement model for the Bayesian update. After introducing the setup and notation, we review the Bayesian update under the Gaussian assumption on the prior (Kalman update), which leads to a regression problem. This interpretation will provide the foundation for developing a nonlinear Bayesian update method in the next section.

\subsection{State variable and measurement setup}
We consider a state variable $U=(u,v)$ in a $d$-dimensional space, where $d = d_1 + d_2$ with $d_1, d_2 \in \mathbb{N}$. The variable $U$ consists of two components: an unobserved component $u$ of dimension $d_1$ and an observed component $v$ of dimension $d_2$. The measurement operation $ H$ acts as a projection of $U$, such that the observed component is given by $v=HU$. The measurement data involves measurement error $\epsilon$, which is Gaussian with zero mean and covariance matrix ${\Gamma}$ in $\mathbb{R}^{d_2}$, $\mathcal{N}(0,{\Gamma})$,
\begin{equation}\label{eq:model:linearprojection}
    {m} = HU+\epsilon=v+\epsilon,\quad\epsilon\sim\mathcal{N}(0,{\Gamma}).
\end{equation}

The above setup with a linear projection measurement has also served as a model in handling nonlinear measurement in the context of ensemble-based Kalman update through an augmented state variable \cite{EKI, EnKFreview}. For a nonlinear measurement $G$ on a state variable $u$ with an additive noise
\begin{equation}\label{eq:model:nonlinearmeasurement}
    {m}=G(u)+\epsilon,
\end{equation}
define the augmented state variable $U$ with the genuine state variable $u$ and an additional component $v=G(u)$, which aligns with the model with a linear projection measurement \eqref{eq:model:linearprojection}.

\subsection{Posterior distribution from Gaussian prior}
The posterior distribution of the state variable $U$ given the measurement ${m}$ is obtained through the Bayes' formula
\begin{equation}\label{eq:Bayes}
p(U|{m})= \frac{p(U)p({m}|U)}{p({m})}.
\end{equation}
With the Gaussian prior for the state variable $U=(u,v)$ and Gaussian measurement noise, the posterior $p(U|{m})$ remains Gaussian. 
When $\tilde{U}=(\tilde{u},\tilde{v})$ and $\tilde{C}=\begin{pmatrix}\tilde{C}_{uu}&\tilde{C}_{uv}\\\tilde{C}_{vu}&\tilde{C}_{vv}\end{pmatrix}$ are the prior mean and covariance, the Kalman update \cite{Kalman} yields the posterior mean $\hat{U}$ and covariance $\hat{C}$:
\begin{eqnarray}
\label{eq:stdKF:meanupdate}\hat{U}&=&\tilde{U}+K({m}-HU)=\tilde{U}+\begin{pmatrix}\tilde{C}_{uv}\\\tilde{C}_{vv}\end{pmatrix}(\tilde{C}_{vv}+{\Gamma})^{-1}({m}-v),\\
\label{eq:stdKF:covupdate}\hat{C}&=&(I-KH)\tilde{C}=\left(I-\begin{pmatrix}\tilde{C}_{uv}\\\tilde{C}_{vv}\end{pmatrix}(\tilde{C}_{vv}+{\Gamma})^{-1}(\tilde{C}_{vu},\tilde{C}_{vv})\right),
\end{eqnarray}
with the Kalman gain defined as
\begin{equation}
K=\tilde{C} H^T(H\tilde{C} H^T+{\Gamma})^{-1}=\begin{pmatrix}\tilde{C}_{uv}\\\tilde{C}_{vv}\end{pmatrix}(\tilde{C}_{vv}+{\Gamma})^{-1}.
\end{equation}
In particular, the updates for the individual components $u$ and $v$ are
\begin{eqnarray}
    \label{eq:postmean:u}\hat{u}&=&\tilde{u}+\tilde{C}_{uv}(\tilde{C}_{vv}+{\Gamma})^{-1}({m}-\tilde{v}),\\
    \label{eq:postmean:v}\hat{v}&=&\tilde{v}+\tilde{C}_{vv}(\tilde{C}_{vv}+{\Gamma})^{-1}({m}-\tilde{v}),
\end{eqnarray}
and the corresponding posterior covariances are
\begin{eqnarray}
    \label{eq:postcov:u}\hat{C}_{uu}&=&I-\tilde{C}_{uv}(\tilde{C}_{vv}+{\Gamma})^{-1}\tilde{C}_{vu},\\
    \label{eq:postcov:v}\hat{C}_{vv}&=&I-\tilde{C}_{vv}(\tilde{C}_{vv}+{\Gamma})^{-1}\tilde{C}_{vv}.
\end{eqnarray}

In order to elucidate the relationship between $u$ and $v$, we proceed from Bayes' formula \eqref{eq:Bayes} as follows:
\begin{equation}\label{eq:splitBayes}
\begin{split}
\frac{p(U)p({m}|U)}{p({m})} &= \frac{p(U)p({m}|v)}{p({m})} \quad\text{(since ${m}$ depends only on $v$)}\\[1mm]
&=\frac{p(u,v)p({m}|v)}{p({m})} \quad\text{(by the definition of joint probability)}\\[1mm]
&=\frac{p(u|v)p(v)p({m}|v)}{p({m})}\\[1mm]
&=p(u|v)\,p(v|{m}).
\end{split}
\end{equation}
This relation holds for any prior distribution. If the prior $p(U)=p(u,v)$ is Gaussian, then $p(v|{m})$ is Gaussian with mean $\hat{v}$ and covariance $\hat{C}_{vv}$ from the Kalman update \eqref{eq:postmean:v} and \eqref{eq:postcov:v}, respectively. Moreover, the conditional distribution $p(u|v)$ is also Gaussian with conditional mean and covariance given by
\begin{equation}\label{eq:conditionalmean}
\mu_{u|v}(v)=\tilde{u}+\tilde{C}_{uv}\tilde{C}_{vv}^{-1}(v-\tilde{v}),
\end{equation}
and
\begin{equation}\label{eq:conditionalcovariance}
C_{u|v}=\tilde{C}_{uu}-\tilde{C}_{uv}\tilde{C}_{vv}^{-1}\tilde{C}_{vu},
\end{equation}
respectively.

Thus, the relation between the posterior means of $u$ and $v$ can be expressed as
\begin{equation}
\hat{u}=\mu_{u|v}(\hat{v}),
\end{equation}
demonstrating that the update of $u$ is obtained through the linear regression of $u$ on $v$ under the Gaussian prior. Note that $\tilde{C}_{uv}\tilde{C}_{vv}^{-1}$ represents the linear regression slope relating $u$ to $v$.

%
%
%
%
%
%
%
%
%
\section{Nonlinear Bayesian Update}\label{sec:NlBU}
In updating $U=(u,v)$, the estimation of $v$ is formulated as a denoising process since $v$ is directly observed (albeit with noise). With high measurement accuracy, the update of $v$ becomes less dependent on its prior distribution. Conversely, updating $u$ proves more challenging due to its potentially nonlinear dependence on $v$. In this section, we propose a nonlinear Bayesian update method that approximates the conditional mean of $u$ given $v$ through a nonlinear regression function, addressing several related issues.

\subsection{Nonlinear regression from the prior ensemble}
A natural extension to handle non-Gaussian priors is to use nonlinear regression to estimate the conditional mean of the unobserved variables given the observed ones. In this approach, various methods—such as polynomial expansions, neural networks, or kernel density estimation (KDE)—can capture the complex relationship between u and v. While previous work (e.g., \cite{hodyss2011ensemble, hodyss2017quadratic}) employed quadratic polynomial expansions, the relationship between u and v may range from linear to highly nonlinear depending on the current state, making a single parametric model potentially biased. Therefore, we adopt a nonparametric regression model to estimate u from v. In particular, we require a regression method that minimizes the need for extensive hyperparameter tuning given the limited size of the prior ensemble. We use a weighted KDE as a robust and practical method for establishing the regression between the observed and unobserved variables, leaving the exploration of alternative techniques for future work.

Let $\{\tilde{U}_k\}=\{\tilde{u}_k,\tilde{v}_k\}$ be the prior ensemble of size $K$. Using two kernels ${\Phi_u}$ and ${\Phi_v}$ for $u$ and $v$ respectively, the prior distribution of $U=(u,v)$ is estimated by
\begin{equation}
    p(u,v)=\frac{1}{K}\sum_{k=1}^{K} {\Phi_v}(v;\tilde{v}_k)   {\Phi_u}(u;\tilde{u}_k).
\end{equation}
From this model, as $\displaystyle p(v)=\int p(u,v)du=\frac{1}{K}\sum_{k=1}^{K} {\Phi_v}(v:\tilde{v}_k)
$, the conditional distribution of $p(u|v)$ is given by
\begin{equation}\label{eq:conditionaluofv}
    p(u|v)=\frac{p(u,v)}{p(v)}=\sum_{k=1}^{K} w_k(v)   {\Phi_u}(u;\tilde{u}_i)
\end{equation}
where $w_k(v)=\frac{{\Phi_v}(v:\tilde{v}_k)}{\sum_{j=1}^{K} {\Phi_v}(v:\tilde{v}_j)}$.
Taking the expected value of $u$ with \eqref{eq:conditionaluofv} gives the Nadaraya–Watson kernel regression \cite{nadaraya1964estimating, watson1964smooth}
\begin{equation}\label{eq:Nadaraya}
\mu_{u|v}(v)=\frac{\sum_{k=1}^{K}{\Phi_v}(v;\tilde{v}_k)\tilde{u}_k}{\sum_{k=1}^K {\Phi_v}(v;\tilde{v}_k)}.
\end{equation}
Then, the the posterior mean of $u$ is estimated by evaluating the nonparametric function for the conditional mean \eqref{eq:Nadaraya} at the posterior mean of $v$, $\hat{v}$.

\subsubsection{Subsampling}
To accurately estimate the conditional mean of $u$ at $v=\hat{v}$, it is crucial to have a sufficient number of ensemble members near $\hat{v}$. A sparse sampling in this region can lead to instability in the KDE estimation due to extrapolation issues \cite{Silverman1986, Scott2015}. To address this, we subsample the ensemble based on the Mahalanobis distance, defined as
\begin{equation}
d_M(v_i, v_j):=\|v_i-v_j\|_{{\Gamma}}=\sqrt{(v_i-v_j)^T{\Gamma}^{-1}(v_i-v_j)}.
\end{equation}
We select only those ensemble members satisfying
\begin{equation}\label{eq:subsampledensemble}
    \mbox{subsampled ensemble}:\{\tilde{u}_{k_m},\tilde{v}_{k_m}\}_{m=1}^{M} \quad\mbox{where}\quad d_M(\tilde{v}_{k_m},\hat{v})\leq 1.
\end{equation}
When the measurement errors are uncorrelated so that ${\Gamma}=\sigma_o^2I$, this criterion is equivalent to selecting ensemble members whose $v$ components lie within an $l_2$ norm ball of radius $\sigma_o$ centered at $\hat{v}$. Once the prior ensemble is subsampled, the nonlinear regression uses only the subsampled ensemble
\begin{equation}\label{eq:regressionwithsubsample}
    \mu_{u|v}(v)=\frac{\sum_{m=1}^{M}{\Phi_v}(v;\tilde{v}_{k_m})\tilde{u}_{k_m}}{\sum_{m=1}^M {\Phi_v}(v;\tilde{v}_{k_m})}.
\end{equation}

The subsampled ensemble size may become very small - or even empty - so that the available prior samples provide little to no information for estimating the unobserved component at $\hat{v}$. In such cases, due to the instability of KDE-based regression when extrapolating beyond observed data, it is recommended to fall back on the linear estimation via the standard Kalman update under the Gaussian assumption.

\subsubsection{Unsupervised clustering for estimation}
In estimating $u$ from the regression function evaluated at $\hat{v}$, we assume a unique functional relationship between $v$ and $u$. However, this assumption may not hold in general. For instance, with a nonlinear measurement such as $v=G(u)$ in \eqref{eq:model:nonlinearmeasurement}, the mapping $G$ might not be invertible, so a unique representation of $u$ in terms of $v$ is not guaranteed. This observation motivates the exploration of alternative estimation strategies—such as mode estimation or clustering-based methods—that take advantage of the complete conditional distribution \eqref{eq:conditionaluofv} at $v=\hat{v}$ rather than relying solely on its mean.

One possible approach is to select the prior ensemble member $\tilde{u}_k$ corresponding to the highest weight $w_k(\hat{v})$. Nevertheless, this strategy depends entirely on the available prior samples and can lead to degraded estimation quality if no sample is sufficiently close to the ground truth. To overcome this limitation, we cluster the conditional distribution and choose the mean of the most densely populated cluster. Without predefined labels or a known number of clusters, we perform unsupervised hierarchical clustering using the SciPy package \cite{scipy2020}. Given the limited size of the prior ensemble (further reduced by subsampling), clustering these samples directly may prove unreliable. Hence, we draw additional samples from the conditional distribution \eqref{eq:conditionaluofv} at $v=\hat{v}$ to enhance the sample density\footnote{We draw many more samples than the original ensemble size since drawing samples from the conditional distribution in the KDE form is fast and the additional samples are used only to calculate the cluster means}. This enriched dataset facilitates a more robust clustering process, after which we compute the mean of each cluster and select the mean of the most probable cluster as the posterior mean of $u$.

\subsubsection{Posteior covariance}
Accurate estimation of the posterior covariance can be investigated using the nonparametric approximation of the conditional distribution. Currently, we do not consider accurate covariance matrix calculations as they often require inflation due to model error or lack of randomness in many applications. In the current study, we set the posterior covariance of $u$ as $\sigma_o^2I$ with no correlation between $u$ and $v$. Thus, the covariance matrix of $U$ is given as 
\begin{equation}\label{eq:postcov}
\hat{C}=\begin{pmatrix}\sigma_o^2I&0\\0&\hat{C}_{vv}\end{pmatrix}
\end{equation}
where $\sigma_o^2$ is the maximxum eigen value of the measurement error covairance ${\Gamma}$.

\subsection{Nonlinear Bayesian Update: Complete Algorithm}
We now describe the complete algorithm for the nonlinear Bayesian update. First, we use an ensemble-based Kalman filter to denoise the observed component $v$, providing the posterior mean $\hat{v}$. Next, we verify that there are enough ensemble members near $\hat{v}$—measured via the Mahalanobis distance—to reliably estimate the conditional mean of $u$. If the number of nearby samples is below the minimum threshold, we revert to the standard ensemble Kalman update. Otherwise, we proceed with the nonlinear update. In this update, if clustering is enabled, additional samples are drawn from the conditional distribution at $v=\hat{v}$ and hierarchical clustering is applied; the mean of the most populated cluster is then used as the refined estimate of $u$. If clustering is not enabled, the nonlinear regression (via the Nadaraya–Watson estimator) is used directly to compute the conditional mean.

Several tuning parameters must be carefully considered. For instance, the threshold for the Mahalanobis distance might need to be adjusted (it may differ from 1), and a minimum number of local samples, $M_{\min}$, must be defined before switching to an ensemble-based filter. In addition, when using hierarchical clustering, choices regarding the flat-cluster formation criterion, the linkage method, and the pairwise distance metric are required. Alternative unsupervised clustering methods (e.g., mean-shift clustering) could also be explored. In Section \ref{sec:experiment}, we discuss our parameter choices, leaving a comprehensive investigation of these variations for future work.
\begin{algorithm}[!bhtp]
\caption{Nonlinear Bayesian Update}
\label{alg:nlBU_revised}
\begin{algorithmic}[1]
\State \textbf{Input:} Prior ensemble $\{(\tilde{u}_k, \tilde{v}_k)\}_{k=1}^K$, measurement ${m}$, measurement error covariance $\Gamma$, clustering flag, minimum subsample size $M_{\min}$
\State \textbf{Output:} Posterior mean $(\hat{u}, \hat{v})$ and posterior ensemble $\{(\hat{u}_k, \hat{v}_k)\}_{k=1}^K$
\State
\State \textbf{Step 1: Denoise $v$ using ensemble Kalman update}
\State Update the prior ensemble $\{\tilde{v}_k\}_{k=1}^{K}$ with ${m}$ to obtain $\{\hat{v}_k\}_{k=1}^{K}$. Compute the posterior mean $\hat{v}$ as the average of $\{\hat{v}_k\}$.
\State
\State \textbf{Step 2: Subsample the Prior Ensemble}
\State Form the subsampled ensemble $\{(\tilde{u}_{k_m}, \tilde{v}_{k_m})\}_{m=1}^{M}$ where $d_M(\tilde{v}_{k_m},\hat{v}) \leq 1$, where $d_M$ is the Mahalanobis distance.
\State
\State \textbf{Step 3: Evaluate Subsample Quality and Determine Update Strategy}
\If{$M < M_{\min}$}
    \State \textbf{(a) Insufficient Local Samples:} Fall back on the standard ensemble Kalman update.
    \State Compute $\hat{u}$ and update $\{\hat{u}_k\}_{k=1}^K$ using the standard Kalman gain.
    \State \textbf{Return:} Posterior mean $(\hat{u},\hat{v})$ and ensemble $\{(\hat{u}_k,\hat{v}_k)\}_{k=1}^K$.
\Else
    \State \textbf{(b) Sufficient Local Samples:} Proceed with the nonlinear update.
\EndIf
\State
\If{clustering is enabled}
    \State \textbf{Step 4a: Hierarchical Clustering for Enhanced Estimation}
    \State i) Generate additional samples from the conditional distribution \eqref{eq:conditionaluofv} at $v=\hat{v}$.
    \State ii) Apply hierarchical clustering to the generated samples.
    \State iii) Select the mean of the most populated cluster as the refined estimate for $\hat{u}$.
\Else
    \State \textbf{Step 4b: KDE-Based Conditional Mean Estimation}
    \State Set 
    \[
        \hat{u} \gets\mu_{u|v}(\hat{v}) = \frac{\sum_{m=1}^{M} \Phi_v(\hat{v};\,\tilde{v}_{k_m})\,\tilde{u}_{k_m}}{\sum_{m=1}^{M} \Phi_v(\hat{v};\,\tilde{v}_{k_m})}\,.
    \]
\EndIf
\State
\State \textbf{Step 5: Compute Posterior Covariance and Generate Posterior Ensemble}
\State For each $k=1,\dots,K$, draw $\eta_k \sim \mathcal{N}(0,\sigma_o^2 I)$ and set
\[
\hat{u}_k \gets \hat{u} + \eta_k\,.
\]
\State \textbf{Return:} Posterior mean $(\hat{u},\hat{v})$ and ensemble $\{(\hat{u}_k, \hat{v}_k)\}_{k=1}^K$.
\end{algorithmic}
\end{algorithm}

%
%
%
%
%
%
%
%
%
\section{Numerical Experiments}\label{sec:experiment}
This section presents several numerical experiments designed to evaluate the performance of the nonlinear Bayesian update method. We focus primarily on low-dimensional problems, which allow the use of relatively large ensemble sizes and eliminate the need for intensive tuning protocols, such as inflation and covariance localization, typically required for high-dimensional settings. The experiments include scenarios involving sequential Bayesian updates (in the context of data assimilation), a PDE-constrained optimization problem (in the context of Ensemble Kalman inversion), and preliminary results for a high-dimensional case, the 40-dimensional Lorenz 96 model.

As a linear update method to be compared with the nonlinear method, we choose the Ensemble adjustment Kalman Filter \cite{EAKF} along with multiplicative inflation and localization if necessary. For KDE, we choose the kernels ${\Phi_u}$ and ${\Phi_v}$ to be Gaussian with covariance matrix estimated from sample covariance scaled by the Scott's rule \cite{Scott2015}. The hierarchical clustering uses the implementation in SciPy package with the single linkage method and the distance criterion for forming flat clusters with threshold value equal to the standard deviation of the prior samples.


The examples considered here include both data assimilation and Ensemble Kalman Inversion (EKI) for solving inverse problems \cite{EKI}. In data assimilation, the statistical accuracy of a random or chaotic dynamical system is improved by incorporating measurement data available at discrete time intervals. Specifically, the system evolves according to
\begin{equation}
U^{n+1}=\Psi(U^n),
\end{equation}
where the dynamical model $\Psi$ propagates the posterior state of $U^n$ forward in time to generate the prior for $U^{n+1}$. When new measurement data becomes available, the Bayesian update is applied to the prior ensemble to obtain the posterior.

Ensemble Kalman Inversion \cite{EKI} applies the ensemble Kalman update to solve the inverse problem
\begin{equation}\label{eq:inverseproblem}
    \operatorname*{argmin}_{u} \|m-G(u)\|^2_{{\Gamma}},
\end{equation}
where the aim is to determine the parameter $u$ that minimizes the weighted misfit between the measurement $m$ and the model prediction $G(u)$. To solve \eqref{eq:inverseproblem}, an augmented state variable $U^n=(u^n,v^n)$ is introduced with the artificial dynamics
\begin{equation}
    U^{n+1}=\Psi(U^n)=(u^n, G(u^n)),
\end{equation}
followed by a Bayesian update using the same measurement data $m$. This process of propagating the state and assimilating measurements is iterated until the estimate converges.

For data assimilation, we evaluate the time-averaged estimation errors over the final 50\% of the assimilation cycles. In the EKI framework, we analyze both the time series of the estimation errors and the convergence speed. The ensemble is initialized from a constant value (typically zero) and perturbed with Gaussian noise having variance $\sigma_{init}^2$.

\subsection{Lorenz 63}
The first example is the Lorenz 63 model \cite{L63}, a three-dimensional system of nonlinear differential equations that exhibits chaotic behavior resembling that of atmospheric dynamics:
\begin{eqnarray}\label{eq:L63}
\frac{dx}{dt} &=& \sigma (y-x),\\[1mm]
\frac{dy}{dt} &=& x(\rho-z) - y,\\[1mm]
\frac{dz}{dt} &=& xy - \beta z.
\end{eqnarray}
We adopt the standard parameter values, $\sigma=10$, $\rho=28$, and $\beta=8/3$. Predictions are obtained by integrating \eqref{eq:L63} using a 4th-order Runge-Kutta method with a time step of $10^{-2}$, which also serves to generate a reference true signal (i.e., there is no model error). Measurements for $y$ are recorded every $\Delta t=0.4$, with a measurement error variance of $\sigma_o^2=10^{-2}$. The ensemble consists of 500 members, initialized at zero and perturbed with Gaussian noise of variance $10^{-1}$. Data assimilation is performed over 500 cycles, with the errors from the final 250 cycles used to assess performance.

For the nonlinear Bayesian update, we consider four combinations: with and without subsampling (SS) and clustering (Cl). Regarding the linear update using EAKF, we use the best results after testing various multiplicative inflation levels ranging from 1 to 1.5 with an increment of 0.05. The nonlinear method shows at least a 8\% error decrease in the worst-case scenario (NlBU with clustering). Individually, subsampling or clustering does not yield significant improvement (in fact, clustering alone worsens the performance compared to the baseline without any enhancements). However, when both clustering and subsampling are applied together, the method achieves the best performance with 17\% and 23\% reductions in the prior and posterior errors, respectively.
\begin{table}[!t]
\centering
\caption{Lorenz 63. Time averaged prior and posterior estimation errors with measurement of $y$}
\begin{tabular}{ccc}
\toprule
& Prior error& Post error \\ \midrule
EAKF&       $2.18\times 10^{-1}$& $1.13\times 10^{-1}$\\
NlBU&           $1.96\times 10^{-1}$& $0.92\times 10^{-1}$\\
NlBU w/ SS&     $1.89\times 10^{-1}$& $0.91\times 10^{-1}$\\
NlBU w/ Cl&     $1.99\times 10^{-1}$& $0.98\times 10^{-1}$\\
NlBU w/ SS Cl&  $1.81\times 10^{-1}$& $0.86\times 10^{-1}$\\
\bottomrule
\end{tabular}
\label{tab:L63y}
\end{table}

\begin{figure}[!bhtp]
\centering
\begin{subfigure}[b]{0.99\textwidth}
    \centering
    \includegraphics[width=\linewidth]{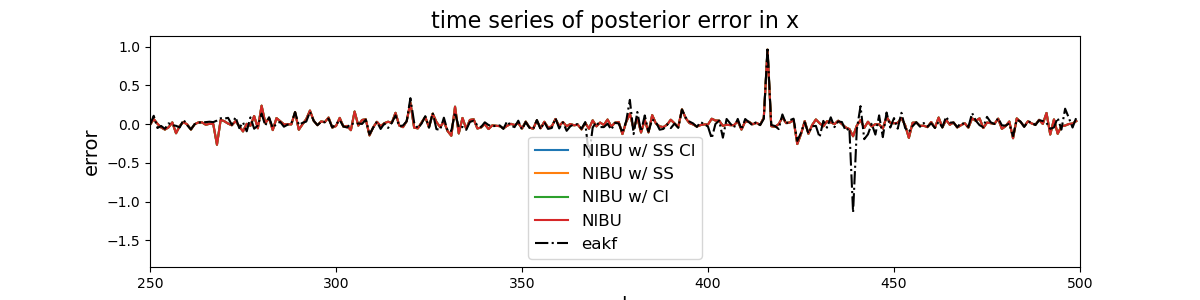}
    \caption{Posterior error of $x$.}
\end{subfigure}
\\[1ex]
\begin{subfigure}[b]{0.99\textwidth}
    \centering
    \includegraphics[width=\linewidth]{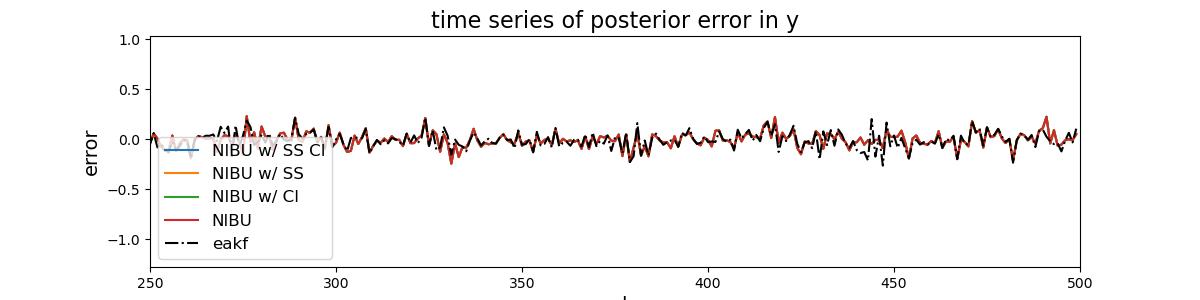}
    \caption{Posterior error of $y$.}
\end{subfigure}
\\[1ex]
\begin{subfigure}[b]{0.99\textwidth}
    \centering
    \includegraphics[width=\linewidth]{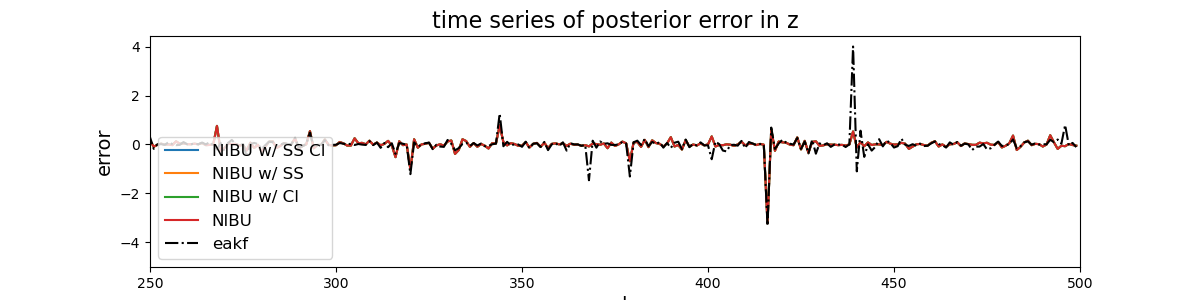}
    \caption{Posterior error of $z$.}
\end{subfigure}
\caption{Time series of the posterior errors of $x$, $y$, and $z$.}\label{fig:L63:ts}
\end{figure}

\subsection{EKI for 2D Darcy flow}
The next experiment is a PDE-constrained optimization problem. The unkonwn variable of interest is the log permeability field $c(x)$ which constitute an elliptic equation related to the subsurface flow described by Darcy flow in the unit square domain $\Omega=(0,1)^2$
\begin{eqnarray}
-\nabla \left(e^{c(x)}\nabla {p}\right)&=&f(x,y), \quad x\in (0,1)^2,\\
{p}(0,y)={p}(1,y)&=&0,\quad y\in (0,1)\\
{p}(x,0)={p}(x,1)&=&\sin(5x),\quad x\in (0,1),
\end{eqnarray}
where $f(x,y)=10\exp\left(-50(x-0.5)^2+50(y-0.5)^2\right)$. The measurement data $m$ is amplitude of the Fourier transform of the pressure field $p(x,y)$ measured at $8\times 8$ uniform grid points inside the domain, yielding a measurement of size 64.

\begin{figure}[htbp]
    \centering
    \begin{subfigure}[b]{0.48\textwidth}
        \centering
        \includegraphics[width=\linewidth]{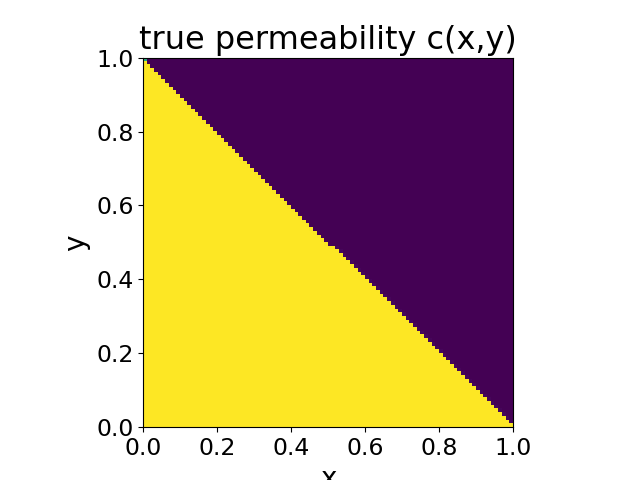}
        \caption{True permeability}
        \label{fig:simpleDarcy:permeability}
    \end{subfigure}
    \hfill
    \begin{subfigure}[b]{0.48\textwidth}
        \centering
        \includegraphics[width=\linewidth]{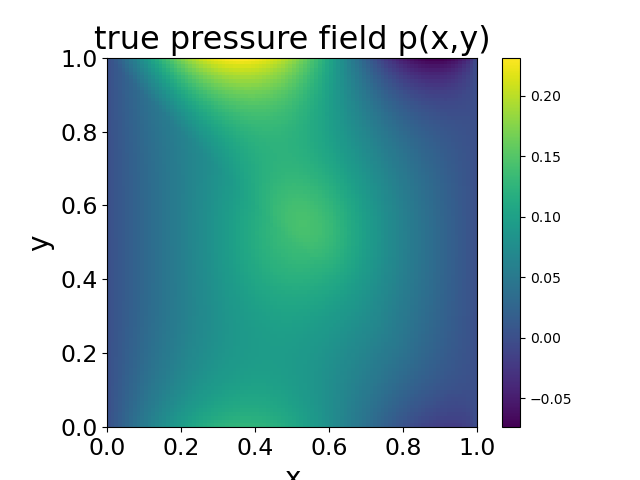}
        \caption{True pressure field}
        \label{fig:simpleDarcy:pressure}
    \end{subfigure}
    \caption{Side-by-side comparison of Darcy flow results.}
    \label{fig:simpleDarcy}
\end{figure}

The permeability field $c(x,y)$ is modeld as discrete yielding only two values in the domain
\begin{equation}
    c(x,y)=\begin{cases}
    u_1, & x+y \ge 1,\\[1mm]
    u_2, & x+y < 1.
    \end{cases}
\end{equation}
Thus, the goal of this inverse problem is to estimate a two-dimensional vector $u=(u_1,u_2)$ from the noisy measurement of the pressure field $m\in\mathbb{R}^{64}$. The true permeability and its corresponding pressure field are shown in Figure \ref{fig:simpleDarcy}.

The measurement noise is assumed to be Gaussian and uncorrelated across measurement components (i.e., ${\Gamma}=\sigma_o^2 I$, with $\sigma_o = 3\times 10^{-3}$). To solve the PDE, we employ the finite element method (FEM) on a $100\times 100$ uniform triangular mesh with a second-order polynomial basis, ensuring convergence of the numerical solver. The ensemble size is set to 1000, with every member initialized at 0.5 and an initial variance of 5.

\begin{figure}[!bhtp]
\centering
\includegraphics[width=.7\textwidth]{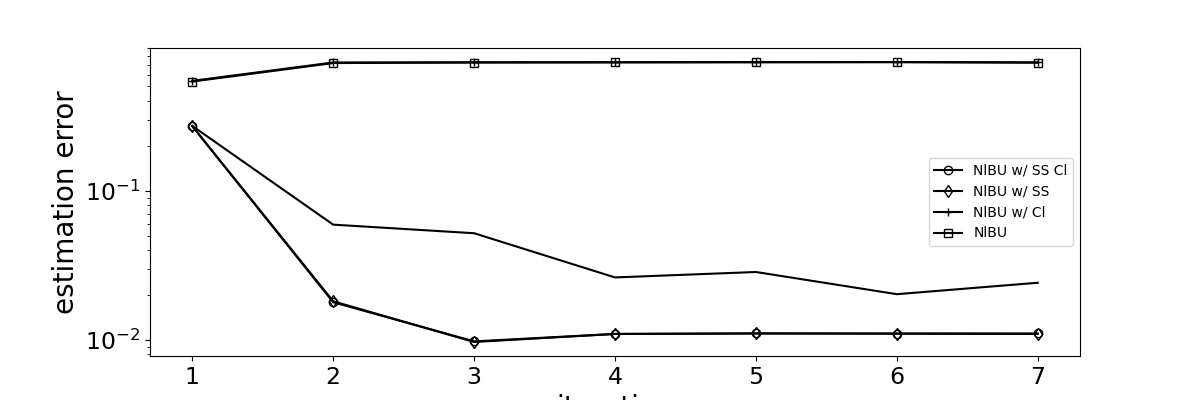}
\caption{estimation error over iteration}\label{fig:simpleDarcy:error}
\end{figure}

Figure \ref{fig:simpleDarcy:error} shows the evolution of the estimation error over iterations. In this experiment, only subsampling makes a significant difference while clustering does not affect the results notably. Without subsampling, the nonlinear method fails to demonstrate any estimation skill, yielding errors on the order of $\mathcal{O}(1)$. When subsampling is applied, the first iteration suffers from an insufficient number of ensemble members (due to the uninformative initialization) and triggers the linear update. Thereafter, the nonlinear update maintains a sufficient number of ensemble samples near the measurement data, converging faster and achieving a smaller estimation error than the linear approach.

\subsection{A high-dimensional case: 40-dimensional Lorenz 96}

The last experiment is a high-dimensional test using the Lorenz 96 model, which is a simple yet widely-used model for turbulent systems and serves as a benchmark for various data assimilation methods. 
In this experiment, we consider a 40-dimensional state defined by
\begin{equation}\label{eq:L96}
\frac{dx_i}{dt}=(x_{i+1}-x_{i-2})x_{i-1}-x_i+F,\quad i=1,\dots,40,
\end{equation}
with periodic boundary conditions (i.e., $x_{-1}=x_{39}$, $x_0=x_{40}$, and $x_{41}=x_1$). 

As in the Lorenz 63 case, the model is integrated using a fourth-order Runge-Kutta method with a time step of $10^{-1}$. Measurements, which are 20-dimensional, are obtained by sampling the even-indexed components $x_{2i}$ (with $i=1,2,\dots,20$) at intervals of $\Delta t=5\times10^{-1}$. The measurement noise is modeled as additive Gaussian noise with covariance $\Gamma = \sigma_o^2 I$, where $\sigma_o^2=10^{-2}$. 
The ensemble comprises 1000 members—25 times the state dimension—with initial states set to zero and perturbed using Gaussian noise of variance $10^{-1}$. Data assimilation is carried out over 500 cycles, and the estimation error is averaged over the final 250 cycles.

\begin{figure}[!bhtp]
\centering
\includegraphics[width=.49\textwidth]{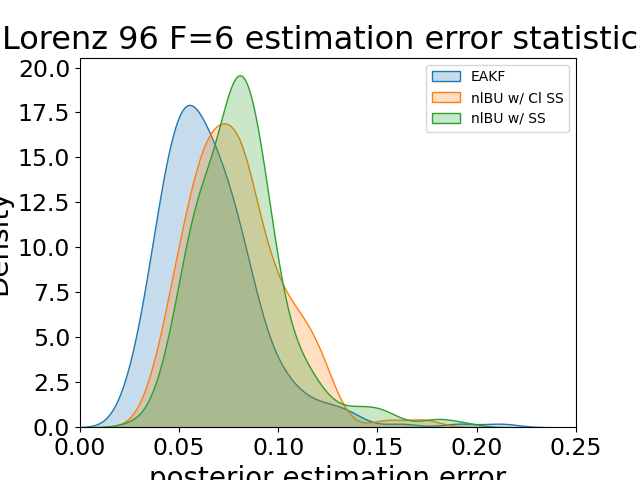}
\includegraphics[width=.49\textwidth]{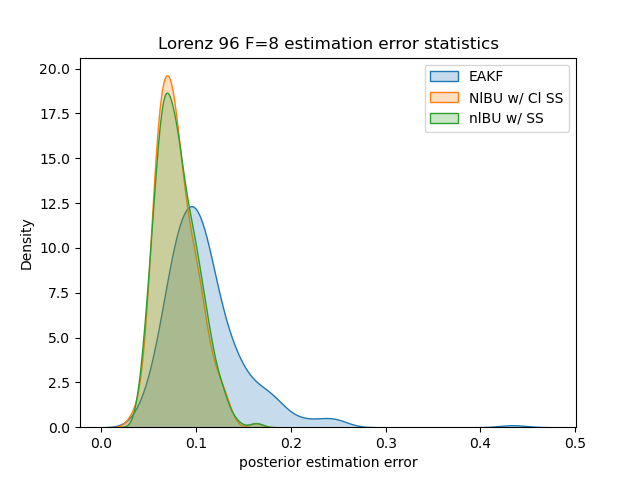}
\caption{Normalized statistical distributions of the posterior error time series for regimes with $F=6$ (left) and $F=8$ (right).} 
\label{fig:L96}
\end{figure}

We examine two regimes by choosing different forcing values: $F=6$ (weakly turbulent) and $F=8$ (moderately turbulent) so as to assess performance under varying degrees of chaos. Table~\ref{tab:L96} reports the time-averaged prior and posterior estimation errors for the linear update (EAKF) and several configurations of the nonlinear update. Distributions of the posterior error time series are shown in Figure~\ref{fig:L96}.  
It is observed that subsampling is crucial for obtaining stable estimation; without subsampling—even when clustering is applied—the nonlinear update suffers from filter divergence. The best performance is achieved when both clustering and subsampling are applied together, although the additional gain from clustering is marginal.

Due to subsampling, the linear update is triggered when the local sample size dropbs below the threshold value 40 (this value is to guaranteed that the sample covariance is not rank defficient). For both test regimes, the linear update is triggered about 80\% of the total assimilation cycles. Even the 20\% application of the nonlinear case can significantly improve the overall estimation skill when $F=8$, dropping the prior and posterior estimation errors by 33\% and 27\%, respectively. 

\begin{table}[!t]
    \centering
    \caption{Lorenz 96 with $F=6$ and $F=8$. Time-averaged prior and posterior estimation errors.} 
    \begin{tabular}{ccc|cc}
        \toprule
         & \multicolumn{2}{c|}{$F=6$}  & \multicolumn{2}{c}{$F=8$} \\
         & Prior error  & Post error  & Prior error  & Post error \\
        \midrule
        EAKF          & $1.09\times 10^{-1}$ & $6.58\times 10^{-2}$ & $2.86\times 10^{-1}$ & $1.09\times 10^{-1}$ \\
        NlBU w/ SS    & $1.35\times 10^{-1}$ & $8.16\times 10^{-2}$ & $1.94\times 10^{-1}$ & $7.98\times 10^{-2}$ \\
        NlBU w/ SS Cl & $1.33\times 10^{-1}$ & $7.88\times 10^{-2}$ & $1.90\times 10^{-1}$ & $7.88\times 10^{-2}$ \\
        \bottomrule
    \end{tabular}
    \label{tab:L96}
\end{table}

%
%
%
%
%
%
%
%
%
\section{Discussions and Conclusions}
In this work, we introduced a nonlinear Bayesian update framework that reformulates the ensemble filtering process as a two‐step update procedure. First, the observed component is denoised using a standard Kalman update, and then the unobserved component is estimated through a nonlinear regression model based on weighted kernel density estimation (KDE). While the regression interpretation is known in the community and several nonlinear regression approaches have been proposed, the current work focuses on addressing key issues such as instability due to extrapolation and the presence of nonunique or multimodal conditional distributions of the unobserved components. These challenges are mitigated by employing local subsampling and unsupervised clustering techniques.

The proposed strategy was evaluated using a variety of test problems, including Lorenz systems and a PDE-constrained optimization problem related to subsurface Darcy flow. Overall, numerical experiments demonstrate that the nonlinear Bayesian update can improve estimation performance over standard ensemble Kalman filtering, particularly in scenarios involving strong nonlinearities and non-Gaussian prior distributions. Future work will explore alternative regression and clustering methods, extend validation to more challenging high-dimensional applications, and investigate strategies for assessing the goodness of fit in the presence of sampling error. Moreover, alternative unsupervised clustering methods, such as mean-shift clustering \cite{Comaniciu2002}, may be incorporated to further refine the conditional estimation.

Regarding the extension to high-dimensional problems, the focus will be on scenarios where the ensemble size is significantly smaller than the state dimension. In ensemble-based Kalman filters for high-dimensional problems, covariance inflation and localization play crucial roles in maintaining filtering performance. Consequently, it is natural to develop comparable techniques within the context of the nonlinear Bayesian update. A recent study has demonstrated that spectral smoothing can alleviate sampling errors due to small ensemble sizes in the data assimilation of turbulent systems \cite{SEM}. Since this approach is independent of the Bayesian update method, incorporating spectral smoothing into the nonlinear Bayesian update represents an interesting direction for future work.

\bibliography{nlBU}
\bibliographystyle{plain}

\end{document}